\def\year{2020}\relax
\documentclass[letterpaper]{article} 
\usepackage{aaai20}  
\usepackage{times}  
\usepackage{helvet} 
\usepackage{courier}  
\usepackage[hyphens]{url}  
\usepackage{graphicx} 
\usepackage{booktabs}
\title{Neural Machine Translation with Byte-Level Subwords}

\author{Changhan Wang$^{\dagger}$, Kyunghyun Cho$^{\dagger\ddagger\star}$ and Jiatao Gu$^{\dagger}$ \\ $^{\dagger}$ Facebook AI Research; $^{\ddagger}$ New York University; $^\star$ CIFAR Global Scholar \\
  {\tt \{changhan, kyunghyuncho, jgu\}@fb.com}}

\begin{document}

\maketitle

\begin{abstract}
Almost all existing machine translation models are built on top of character-based vocabularies: characters, subwords or words. Rare characters from noisy text or character-rich languages such as Japanese and Chinese however can unnecessarily take up vocabulary slots and limit its compactness. Representing text at the level of bytes and using the 256 byte set as vocabulary is a potential solution to this issue. High computational cost has however prevented it from being widely deployed or used in practice. In this paper, we investigate byte-level subwords, specifically byte-level BPE (BBPE), which is compacter than character vocabulary and has no out-of-vocabulary tokens, but is more efficient than using pure bytes only is. We claim that contextualizing BBPE embeddings is necessary, which can be implemented by a convolutional or recurrent layer. Our experiments show that BBPE has comparable performance to BPE while its size is only $1/8$ of that for BPE. In the multilingual setting, BBPE maximizes vocabulary sharing across many languages and achieves better translation quality. Moreover, we show that BBPE enables transferring models between languages with non-overlapping character sets.
\end{abstract}

\section{Introduction}
It has become a standard practice to build a vocabulary in neural machine translation (NMT)~\cite{bahdanau2014neural,sutskever2014sequence} using byte-pair encoding (BPE)~\cite{sennrich:acl2015}. In this practice, we notice that BPE is used at the level of characters rather than at the level of bytes, which is more common in data compression. We suspect this is because text is often represented naturally as a sequence of characters, although it has recently been noticed that byte representation of text has its own advantages, such as compactness (up to 256 possible values) and being agnostic to languages.

In this paper, we look into byte-level ``subwords" that are used to tokenize text into variable-length byte $n$-grams, as opposed to character-level subwords in which we represent text as a sequence of character $n$-grams. We specifically focus on byte-level BPE (BBPE), examining compact BBPE vocabularies in both bilingual and multilingual settings as well as in a novel setup of transfer learning to a new language with a non-overlapping character set.

\begin{figure}[t]
    \centering
    \small
    \includegraphics[width=0.42\textwidth]{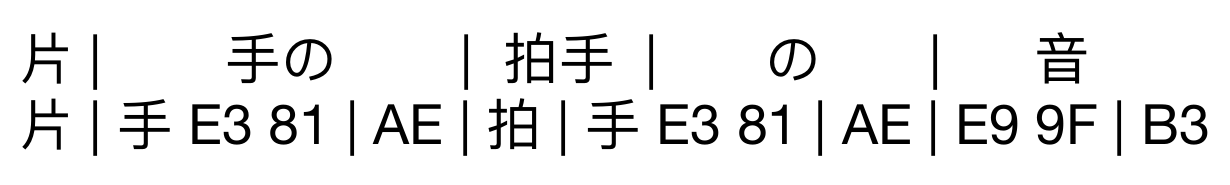}
    \caption{BPE (upper) and BBPE (lower) tokenization of a Japanese sentence. Bytes (from partial characters) are represented by hexadecimal digits.}
    \label{fig:bbpe_tokenization}
\end{figure}

\section{Byte Level Text Representation}

\paragraph{Encoding Byte-Level Representation}
We consider UTF-8 encoding of text, which encodes each Unicode character into 1 to 4 bytes. This allows us to model a sentence as a sequence of bytes instead of characters. While there are 138K Unicode characters covering over 150 languages, we represent a sentence in any language as a sequence of UTF-8 bytes (248 out of 256 possible bytes).

A byte sequence representation of text is often much longer (up to 4x) than a character sequence representation, which makes it computationally demanding to use bytes as they are. As an alternative, we consider segmenting a byte sequence into variable-length $n$-grams (byte-level ``subwords"). Specifically, we learn BPE vocabulary on the byte-level representation which extends UTF-8 byte set with byte $n$-grams. We denote this type of vocabulary as B(yte-level)BPE in the rest of the paper. Figure \ref{fig:bbpe_tokenization} shows an example of BBPE tokenization.

BBPE symbols can be partial characters shared by different characters or the combination of complete and partial characters. This arbitrariness may necessitate incorporating a larger context surrounding each symbol for disambiguation and learning the character boundaries. In this work, we base our experiments on Transformer~\cite{vaswani:nips2017} models. We propose to use either a depth-wise convolutional layer \cite{kaiser2017depthwise} or a bidirectional recurrent layer with gated recurrent units~\cite[GRU,]{cho2014learning} to contextualize BBPE embeddings before feeding them into the model:

$$\mathbf{x}_{ctx\_emb} = \textrm{DepthWiseConv}(\mathbf{x}_{emb})$$
or
$$\mathbf{x}_{ctx\_emb} = \textrm{BiGRU}(\mathbf{x}_{emb})$$

\paragraph{Decoding with Byte-Level Subwords}
While any sentence can be represented as a byte sequence, the converse is, however, not necessarily true in that there are byte sequences that do not translate to valid character sequences. Empirically, we find that invalid outputs from trained models are very rare. We do not observe any in the experiments described below (note that one of them does have a large test set of 165K examples). And a common error pattern in half-trained models is redundant repeating bytes. In our system, we try to recover as many Unicode characters as possible from this error pattern efficiently in linear time. The algorithm is as follows: For a given byte sequence $\{B\}_{k=1}^N$, we denote the maximum number of characters that we can recover from it as $f(k)$. Then $f(k)$ has optimal substructure and can be solved by dynamic programming:
\begin{equation}
f(k) = \max_{t=1,2,3,4}\{f(k-t) + g(k - t + 1, k)\}
\label{eq:dp_recovery}
\end{equation}
where $g(i, j) = 1$ if $\{B\}_{k=i}^j$ corresponds to a valid character, otherwise $0$. When $f(k)$ is calculated recursively, we also record the selections at each position $k$ so that we can recover the solution through backtracking. The design of UTF-8 encoding ensures the uniqueness of this recovery process: for a character UTF-8 encoded with multiple bytes, its trailing bytes will not make a valid UTF-8 encoded character. Then the best selection in Eq. \ref{eq:dp_recovery} is unique and so is the final solution.

\section{Experimental Settings}

\paragraph{Datasets}
We run experiments on three bilingual corpora as well as a many-to-English multilingual dataset:

\begin{itemize}
\item English-German (En-De): we replicate the same setting of~\cite{vaswani:nips2017} which uses WMT 2014  \footnote{\url{http://statmt.org/wmt14/translation-task.html}} 
data (newstest13 for validation and newstest14 for testing)
\item Japanese-English (Ja-En): we follow~\cite{michel2018mtnt} and concatenate KFTT\footnote{http://www.phontron.com/kftt}~\cite{neubig11kftt}, TED\footnote{https://wit3.fbk.eu/mt.php?release=2017-01-trnted}~\cite{cettolo2012wit3} and JESC\footnote{https://nlp.stanford.edu/projects/jesc}~\cite{pryzant_jesc_2017} to construct training, validation and test sets.
\item Sinhala-English (Si-En): we use the data from FLoRes \cite{guzan:arxiv2019}.
\item Many-to-English (X-En): we adopt the TED Talks corpus complied by \cite{Ye2018WordEmbeddings}, which includes parallel data for 59 languages.  For our experiments, we use English as target and the other 58 languages as source. We sample 22K examples from the 135K development set for validation.
\end{itemize}
Table \ref{tab:dataset_volume} shows an overview statistics of these datasets. We learn (B)BPE vocabularies jointly on source and target sentences using SentencePiece~\cite{sentencepiece}.

\begin{table}[h]
    \centering
    \begin{tabular}{c|cccc}
     & En-De & Ja-En & Si-En & X-En \\
    \toprule
    Train & 4.5M & 3.5M & 405K & 5.1M \\
    Dev & 3K & 4K & 3K & 22K$^\star$ \\
    Test & 3K & 12K & 3K & 165K \\
    \end{tabular}
    \caption{Dataset statistics in number of sentences. $\star$ Sub-sampled from the full 135K development set.}
    \label{tab:dataset_volume}
\end{table}

\begin{table}[h]
    \centering
    \begin{tabular}{c|cccccc}
    Model & $N$ & $d_{model}$ & $d_{ff}$ & $h$ & $P_{drop}$ & Params \\
    \toprule
    $T_{flores}$ & 5 & 512 & 2048 & 2 & 0.4 & 38M \\
    $T_{base}$ & 6 & 512 & 2048 & 8 & 0.1 & 44M \\
    $T_{big}$ & 6 & 1024 & 4096 & 16 & 0.3 & 180M \\
    \end{tabular}
    \caption{Transformer models used in the experiments (using the notations in Vaswani et al. 2017).}
    \label{tab:table_models}
\end{table}

\paragraph{Models and Learning}
We use Fairseq~\cite{ott2019fairseq} to train Transformers~ \cite{vaswani:nips2017} with the same learning rate schedule in the original paper. All model configurations are listed in table \ref{tab:table_models}. We set attention and ReLU dropout to $0.1$, except Si-En for which we use $0.2$. We use 0.2 residual dropout for $T_{base}$ models in X-En. We use a kernel size of 5 and a padding of 2 on both sides for all convolutional layers.

\paragraph{Inference and Evaluation} 
We set beam width to 4 for En-De and 5 for the other and use the best checkpoint by validation loss to generate the predictions. We calculate case-sensitive tokenized BLEU ~\cite{papineni2002bleu} as the metrics using sacreBLEU~\cite{sacrebleu}.

\section{Results and Analysis}
\subsection{Qualitative Comparison: BPE vs. BBPE}

\paragraph{Symbol Frequency Distribution}

\begin{figure*}[t]
    \centering
    \includegraphics[width=0.48\textwidth]{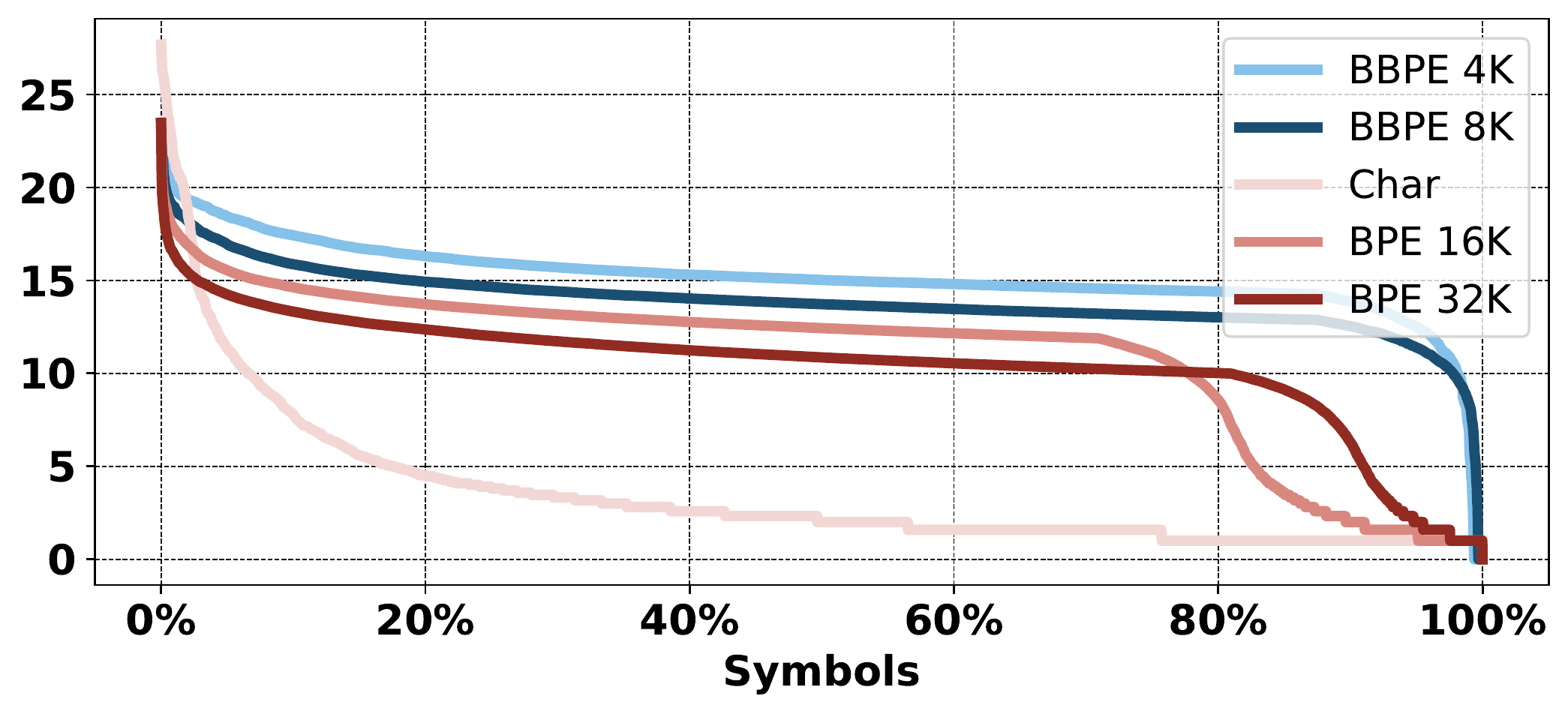}
    \includegraphics[width=0.48\textwidth]{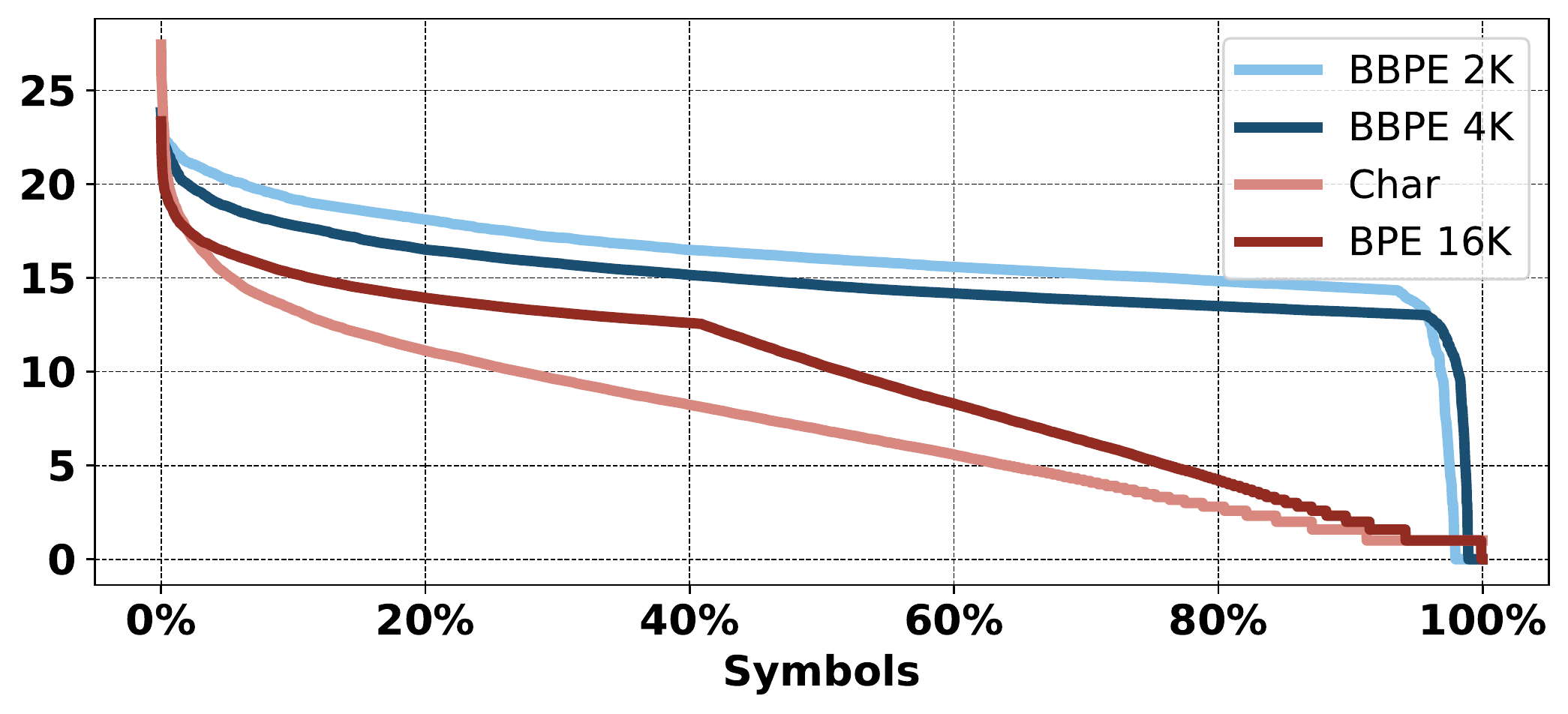}
    \caption{Symbol frequencies (in log2 scale) for En-De (left) and X-En (right) vocabularies. BBPE enables a more consistent distribution of vocabulary across frequencies.}
    \label{fig:symbol_freq}
\end{figure*}

\begin{figure*}[t]
    \centering
    \includegraphics[width=\linewidth]{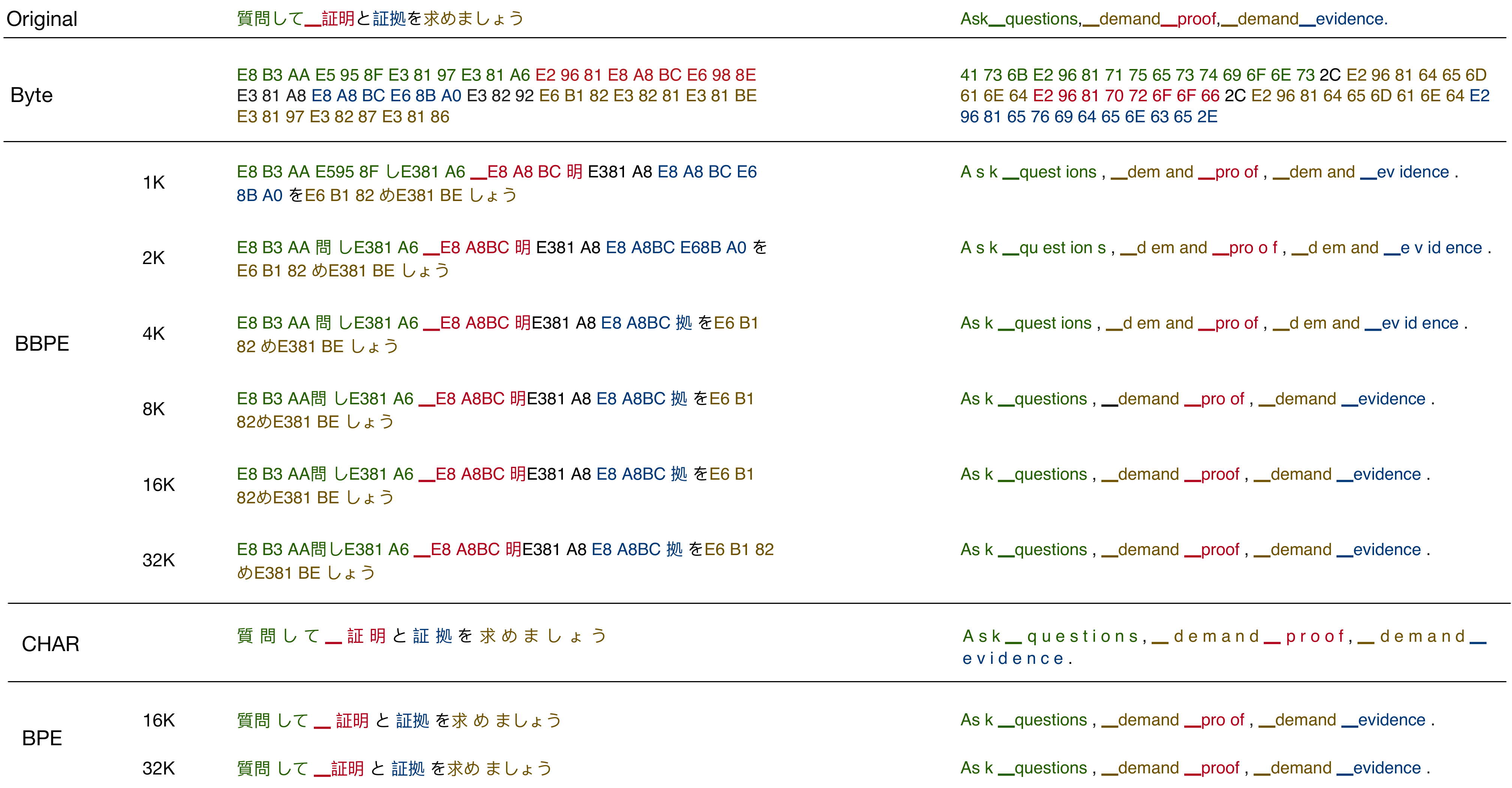}
        \caption{An example from Ja-En tokenized with different vocabularies. Raw spaces are replaced by underscores and spaces are used to split tokens. We can observe how tokens look like as the tokenization granularity goes from fine to coarse: Byte (256) $\rightarrow$ BBPE (1K, 2K, 4K, 8K) $\rightarrow$ Char (8K) $\rightarrow$ BBPE (16K, 32K) $\rightarrow$ BPE (16K, 32K).}
    \label{fig:ja_en_tokenization}
\end{figure*}

Since the construction of BBPE vocabulary starts from UTF-8 byte set, it has the flexibility of decomposing rare characters into byte $n$-grams from the vocabulary instead of including them directly. This frees vocabulary slots for other frequent symbols. Figure \ref{fig:symbol_freq} compares the symbol frequency distribution of BPE and BBPE. We can see that BBPE symbols are more evenly distributed than BPE ones, even when the latter has already been much more evenly distributed than pure characters. By setting different BBPE vocabulary sizes, we can control the level of rare character decomposition and symbol sharing across different characters. Table~\ref{tab:partial_char} shows the ratio of BBPE tokens with partial characters. We can see that large portion of rare characters are decomposed on Ja-En and X-En, which has a large character set of 8K and 11K, respectively. Figure~\ref{fig:ja_en_tokenization} provides an example from Ja-En tokenized with different BBPE vocabularies, where we can see how tokens look like as the tokenization granularity goes from fine to coarse.

\begin{table}[h]
    \centering
    \begin{tabular}{c|ccccc}
        BBPE & 2K & 4K & 8K & 16K & 32K \\
        \toprule
        En-De & 4.3\% & 4.9\% & 5.5\% & 6.1\% & 6.5\% \\
        Ja-En & 46.0\% & 47.6\% & 49.4\% & 51.2\% & 34.8\% \\
        X-En & 36.8\% & 39.1\% & 41.3\% & 43.6\% & 23.0\% \\
    \end{tabular}
    \caption{Ratio of BBPE tokens with partial characters.}
    \label{tab:partial_char}
\end{table}

\paragraph{Cross-Lingual Sharing}

\begin{figure}[t]
    \centering
    \includegraphics[width=0.47\textwidth]{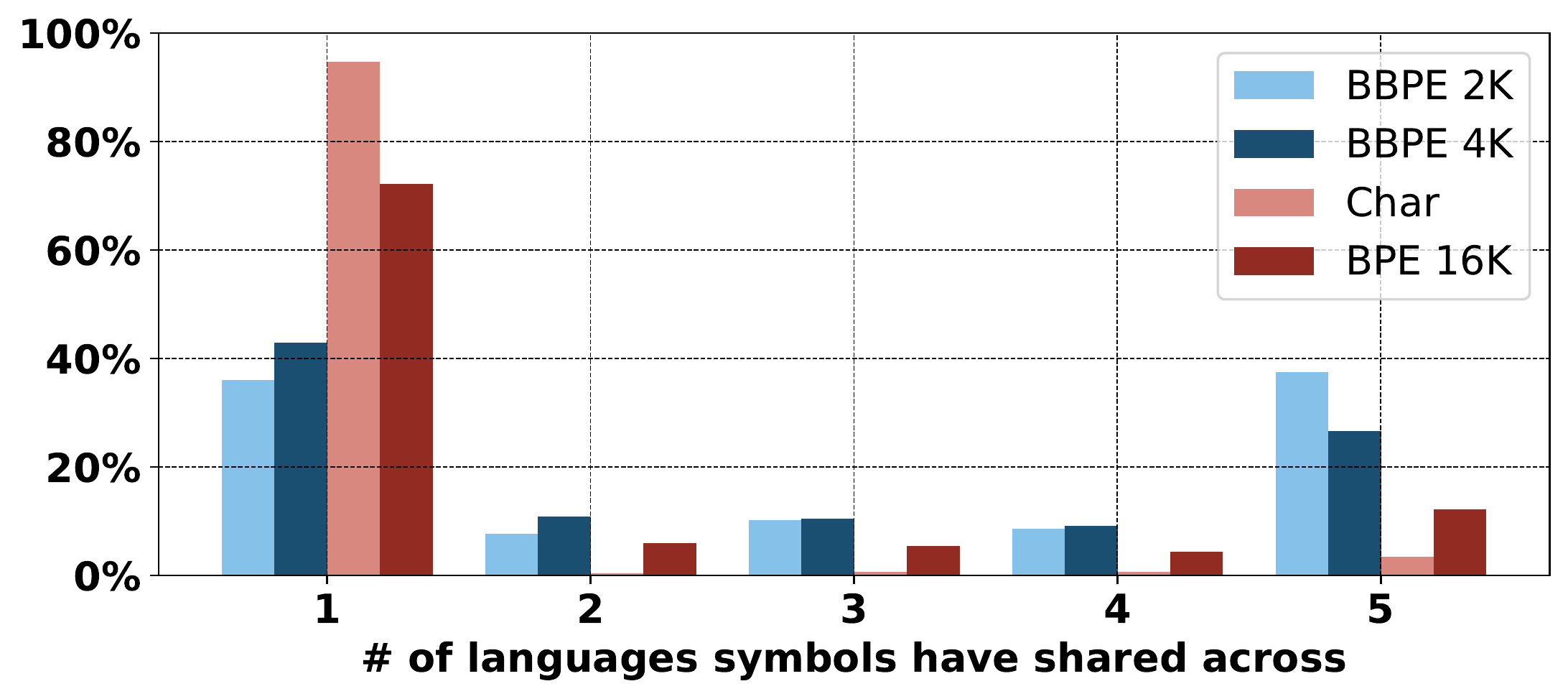}
    \caption{The numbers of languages symbols have shared across Ar, He, Ru, Ko and It (from X-En). Note that these languages have mutually different character sets.}
    \label{fig:symbol_sharing}
\end{figure}

In the multilingual setting, symbol sharing also happens across different languages despite the different writing systems. This allows maximizing parameter sharing not only for the model part but also the vocabulary part in a universal model. Figure \ref{fig:symbol_sharing} illustrates the level of BBPE symbol sharing across the top 5 languages (by number of train examples) in X-En whose writing systems are different from each other.

\paragraph{Impact on Sequence Lengths}

Compared to BPE, BBPE symbols are generally finer-grained with shorter byte-level lengths, which results in longer tokenized sequences as well as longer training and inference time. BBPE, however, is optimized for compression-based objective (the same as BPE), and is still more efficient than character vocabulary. Table \ref{tab:table_sentence_len} lists the average lengths of training sentences tokenized with different vocabularies. We can observe that sentences tokenized with BBPE have significantly shorter lengths than the character ones, even when the BBPE vocabulary is much smaller (for example only $1/5$ of character set size on X-En). Another observation is that source-target length ratio for BBPE tends to be much larger when source character set and target character set have very different sizes (for example 11K for X-En source side and 0.1K for the target side). And this situation becomes more severe when BBPE vocabulary size increases. In this case, alignments may be more difficult to learn during model training, since target tokens need attentions on multiple source tokens more often.

\begin{table*}[t]
    \centering
    \begin{tabular}{cc|c|cccccccc|ccc|ccc}
     &  & Byte & \multicolumn{8}{c|}{BBPE} & \multicolumn{3}{c|}{Char} & \multicolumn{3}{c}{BPE} \\
    & & 256 & 1K & 2K & 3K & 4K & 8K & 11K & 16K & 32K & 3K & 8K & 11K & 8K & 16K & 32K \\
    \toprule
    En-De & Source & 143 & 57 & 48 & 43 & 41 & 36 & & 33 & 30 & 143 & & & 40 & 33 & 31 \\
     & Target & 160 & 64 & 55 & 50 & 48 & 42 & & 38 & 35 & 157 & & & 43 & 36 & 32 \\
    \midrule
    Ja-En & Source & 55 & 28 & 26 & & 24 & 23 & & 21 & 21 & & 19 & & & 12 & 10 \\
     & Target & 53 & 23 & 20 & & 17 & 15 & & 15 & 13 & & 52 & & & 15 & 14 \\
    \midrule
     X-En & Source & 126 & 77 & 70 & & 65 & 62 & 60 & 59 & 57 & & & 89 & & 40 & 32 \\
    & Target & 103 & 49 & 43 & & 37 & 33 & 32 & 30 & 27 & & & 103 & & 35 & 30 \\
    \end{tabular}
    \caption{Average lengths of training sentences tokenized with different vocabularies.}
    \label{tab:table_sentence_len}
\end{table*}

\subsection{Importance of Contextualization}

\begin{figure}[t]
    \centering
    \includegraphics[width=0.48\textwidth]{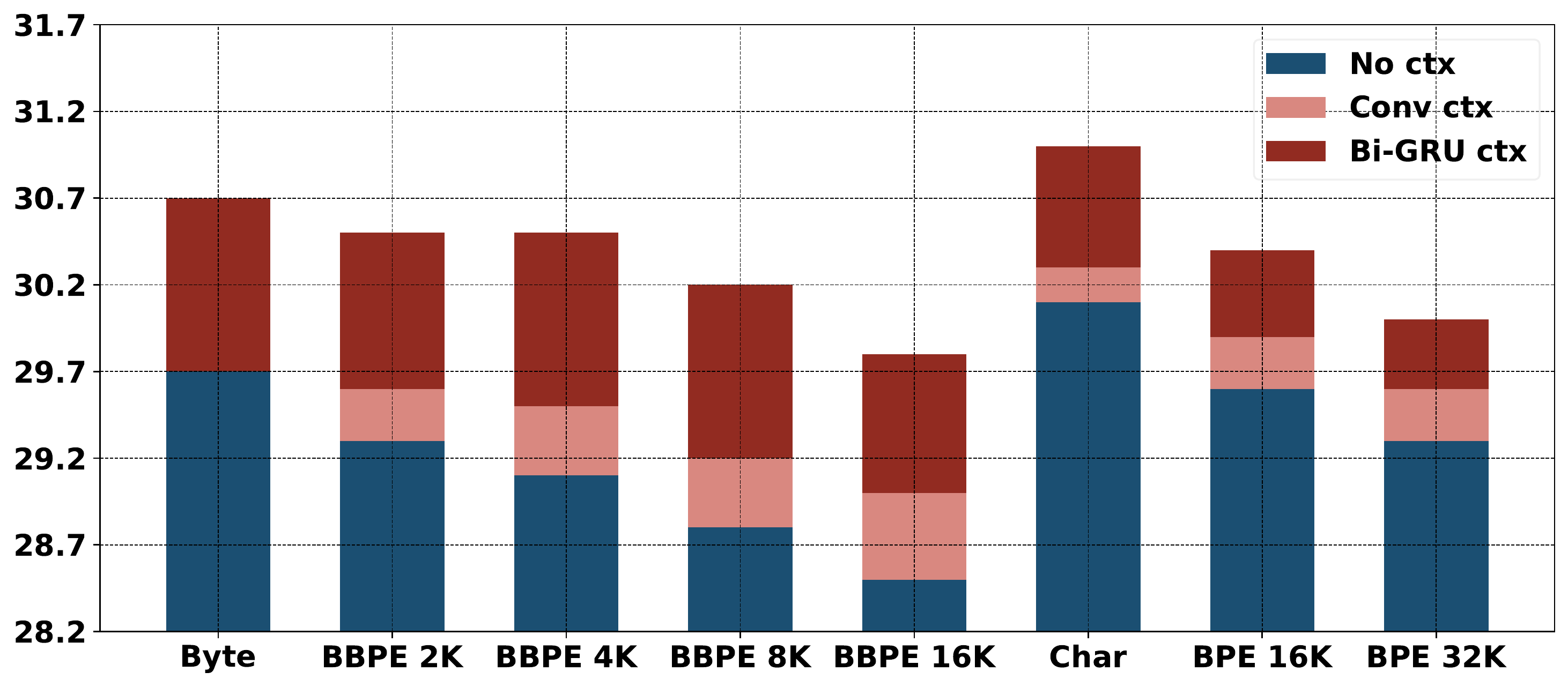}
    \caption{X-En validation BLEU for models without contextualization, with local contextualization (depth-wise convolution) and with long-range contextualization (Bi-GRU). The y axis starts from 28.2 to focus on the gain portions and facilitate comparison across different vocabularies.}
    \label{fig:ctx_gain}
\end{figure}

We compare three different ways of contextualizing token embeddings; none, 1-layer convolution and 1-layer bi-GRU, on X-En with $T_{base}$ model. We observe from Figure~\ref{fig:ctx_gain} that all kinds of vocabularies can benefit from embedding contextualization. Performance gains are more significant on fine-grained vocabularies: byte, character and BBPE. For BBPE, long-range contextual information from Bi-GRU brings over $4\%$ gain on validation BLEU in all the cases. Encoding context in the token embeddings reduces the difficulties of learning attentions on multiple source tokens and makes model training easier. In the following experiments, we contextualize BBPE with Bi-GRU by default. We denote (B)BPE with Bi-GRU as ``(B)BPE $<$size$>$+" and the one without contextualization as ``(B)BPE $<$size$>$". And we similarly define ``Byte+" and ``Char+".

\subsection{BBPE on Noisy Character Sets}

The En-De training set has quite a few noisy sentence pairs often containing a few non-latin alphabets due to misalignment and code-switched sentences. This leads to a 3.4K character set, while in contrast, English and German both have less than 30 alphabets. Since BPE includes all characters, those rare characters will waste quite a lot of BPE vocabulary slots. For comparison, we try with small BBPE 2K and 4K vocabulary where rare characters are excluded. We find that their performance are comparable to the BPE 32K baseline while having smaller model capacity (see table \ref{tab:table_en_de}).

\begin{table}[h]
    \centering
    \begin{tabular}{ll|c|c}
     &  & Test BLEU & Params \\
    \toprule
    $T_{base}$ & Byte+ & 26.59 & 45M \\
    & BBPE 2K+ & 26.98 & 47M \\
    & BBPE 4K+ & 27.08 & 47M \\
    & Char+ & 26.73 & 47M \\
    & BPE 32K & 27.31 & 61M \\
    & BPE 32K+ & \textbf{27.41} & 62M \\
    & BPE 37K$^\star$ & 27.3 & 65M \\
    \midrule
    $T_{big}$& Byte+ & 26.94 & 181M \\
    & BBPE 2K+ & \textbf{28.78} & 183M \\
    & BBPE 4K+ & 28.27 & 185M \\
    & Char+ & 27.24 & 185M \\
    & BPE 32K & 28.36 & 210M \\
    & BPE 32K+ & 28.77 & 215M \\
    & BPE 37K$^\star$ & 28.4 & 213M \\
    \end{tabular}
    \caption{En-De test BLEU. $\star$ \cite{vaswani:nips2017}.}
    \label{tab:table_en_de}
\end{table}

\subsection{BBPE on Character-Rich Languages}

Languages using logographic writing systems, such as Chinese and Japanese, can have over 50K characters, while only a small portion of them are frequently used. Our Ja-En dataset has a set of 7.9K characters, where 99.99\% tokens in the training set are covered by the top 2.4K characters. With this observation, we experiment with BBPE 4K which is roughly $50\%$ of the character set size. We find that BBPE is comparable to BPE and even outperforms BPE when using larger $T_{big}$ model (see table \ref{tab:table_ja_en}).

\begin{table}[h]
    \centering
    \begin{tabular}{ll|ccc|c}
    & & KFTT & TED & JESC & All \\
    \midrule
    \multicolumn{2}{c|}{\# of train samples} & 440K & 223K & 2.8M & 3.5M \\
    \multicolumn{2}{c|}{\# of test samples} & 1.2K & 8.5K & 2K & 11.7K \\
    \toprule
    \multicolumn{2}{c|}{Michel et.al. (2018)} & 20.77 & 13.25 & 18.00 & - \\
    \midrule
    $T_{base}$& Byte+ & 23.12 & 15.14 & 15.69 & 16.27 \\
    & BBPE 4K+ & \textbf{24.15} & 15.59 & 16.10 & 16.80 \\
    & Char+ & 23.67 & 15.26 & 15.68 & 16.43 \\
    & BPE 16K+ & 23.63 & \textbf{16.15} & \textbf{16.18} & \textbf{17.19} \\
    \midrule
    $T_{big}$& Byte+ & 23.68 & 16.08 & 16.29 & 17.46 \\
    & BBPE 4K+ & 23.88 & \textbf{19.0} & \textbf{17.93} & \textbf{19.58} \\
    & Char+ & 23.71 & 16.69 & 17.01 & 18.33 \\
    & BPE 16K+ & \textbf{24.08} & 18.34 & 17.89 & 19.14 \\
    \end{tabular}
    \caption{Ja-En test BLEU scores.}
    \label{tab:table_ja_en}
\end{table}

\subsection{BBPE on Many-to-En Translation}

Our many-to-En dataset contains 58 languages (parallelly to English) and 10.8K characters from different writing systems, between which characters are not necessarily shared. The characters, however, share byte $n$-grams. We experiment with BBPE 2K and 4K that have $12.5\%$ and $25\%$ size of the baseline BPE vocabulary. As shown in Table \ref{tab:x_en}, both of them beat the BPE baseline on overall BLEU as well as on most of the languages both with high and low resources (Note that the test set is as large as 165K and even small gaps in BLEU may suggest significant difference). We also notice that byte model and character model perform significantly better than all BPE and BBPE models in this multilingual setting. This might be because that BBPE and BPE suffer from imbalanced source and target sentence lengths as well as various token granularities in multilingual parallel sentences (sources in different languages and granularities into same targets). Nonetheless, BBPE is still the most practical solution since it makes a good balance between performance (better BLEU than BPE) and speed (much shorter tokenized sentences than characters and bytes).

\begin{table*}[t]
    \centering
    \begin{tabular}{cc|cccc|cccc|c|c}
     & & Ar & De & He & It & Az & Be & Gl & Sk & All & Params \\
     \midrule
    \multicolumn{2}{c|}{\# of train examples} & 213K & 167K & 211K & 203K & 5.9K & 4.5K & 10K & 61K & 5.1M & \\
    \multicolumn{2}{c|}{\# of test examples} & 6K & 4.5K & 5.5K & 5.6K & 0.9K & 0.7K & 1K & 2.4K & 165K & \\
    \toprule
    \multicolumn{2}{c|}{Aharoni et al. 19} & 25.93 & 28.87 & 30.19 & 32.42 &  &  &  &  & & \\
    \multicolumn{2}{c|}{Neubig \& Hu 18} & & & & & 11.7 & 18.3 & 29.1 & 28.3 & & \\
    \midrule
    $T_{base}$ & Byte+ & 31.13 & 35.98 & 36.77 & 38.36 & 14.64 & \textbf{25.12} & 35.12 & 33.08 & 30.38 & 45M \\
    & Char+ & \textbf{31.52} & \textbf{36.73} & \textbf{36.85} & \textbf{38.62} & \textbf{15.40} & 24.90 & \textbf{35.44} & \textbf{33.31} & \textbf{30.75} & 51M \\
    \midrule
    $T_{base}$ & BBPE 2K+ & \textbf{30.79} & \textbf{35.53} & \textbf{36.27} & \textbf{37.82} & 13.64 & 24.70 & \textbf{34.17} & \textbf{32.83} & \textbf{29.91} & 46M \\
    & BBPE 4K+ & 30.64 & 34.93 & 36.07 & 37.62 & \textbf{13.76} & \textbf{24.84} & 33.90 & 32.12 & 29.74 & 47M \\
    & BPE 16K & 29.70 & 34.35 & 34.47 & 37.02 & 13.28 & 24.61 & 33.55 & 31.72 & 29.00 & 53M \\
    & BPE 16K+ & 30.20 & 34.97 & 35.55 & 37.49 & 12.65 & 23.66 & 33.95 & 32.16 & 29.62 & 54M \\
    & BPE 32K & 29.02 & 34.08 & 34.18 & 36.63 & 12.56 & 22.48 & 32.33 & 31.26 & 28.81 & 61M \\
    & BPE 32K+ & 29.87 & 34.64 & 35.26 & 37.43 & 12.35 & 22.05 & 33.62 & 31.61 & 29.43 & 62M \\
    \end{tabular}

    \caption{X-En test BLEU on all 58 languages, top-4 (Ar, De, He, It) and bottom-4 (Az, Be, Gl, Sk) languages by number of training samples. Note that the test set is very large (165K) and even small gaps in BLEU may suggest significant difference.}
    \label{tab:x_en}
\end{table*}
\begin{table}
    \centering
    \begin{tabular}{ll|cc|c}
    & & Train & Finetune & BLEU \\
    \toprule
    $T_{flores}$& BPE 5K$^\star$ & Si-En & - & 7.2 \\
    & BBPE 4K+ & Si-En & - & 7.1 \\
    \midrule
    $T_{flores}$ & BBPE 4K+ & X-En & - & 0.3 \\
    &BBPE 4K+ & X-En & enc & 8.3 \\
    &BBPE 4K+ & X-En & enc, dec & 8.1 \\
    &BBPE 4K+ & X-En & embed, enc & \textbf{9.0} \\
    &BBPE 4K+ & X-En & all & 8.6 \\
    \end{tabular}
    \caption{Transferring pretrained X-En model to Si-En. BBPE 4K is learned on X-En. $\star$  \cite{guzan:arxiv2019}.}
    \label{tab:si_en}
\end{table}

\subsection{Transfer Learning on Unseen Characters}

Because BBPE contains all UTF-8 bytes and has no out-of-vocabulary tokens, BBPE-based models can be transferred between languages with non-overlapping character sets. In comparison, it is impossible to do so with character-based vocabularies without replacing the vocabulary and re-training embeddings from scratch. Our Si-En dataset has 77 Sinhala scripts that are disjoint with the X-En character set. We experiment transferring a pretrained (on X-En) BBPE 4K $T_{flores}$ model to this dataset while reusing the original vocabulary. As shown in table \ref{tab:si_en}, the transferred model gains 0.9-1.8 BLEU points compared to the baselines, suggesting the generality of pretrained BBPE embeddings and its ability to adapt to different languages with unseen characters. This transfer learning paradigm is free from the limitation of out-of-vocabulary tokens and can be very generic. We just show the extreme case of totally unseen character set, but the pre-trained model may also be transferred to any languages and datasets to improve performance or warm-start model training to save time.

\section{Related Work}

\paragraph{Subword Vocabularies}
Previous works have shown that finer-grained vocabularies consistently outperforms word-level vocabularies in many settings, for example, vocabularies based on morpheme segmentation~\cite{niessen2000improving,luong2013better}, byte-pair encoding~\cite{sennrich:acl2015} and vocabularies from unigram language model~\cite{kudo2018subword}. Our byte-level subword vocabularies are based on byte-pair encoding, while we use bytes as the basic units to compose subwords.

\paragraph{Character Vocabulary}
Existing works also explored pure character vocabulary for machine translation.~\cite{kim2016character} proposed building word representations from character ones;~\cite{chung2016character} removed the restriction of word boundaries and directly learned decoding in character level;~\cite{lee2017fully} further extended it to a fully character-level model in a multilingual setting;~\cite{cherry2018revisiting} showed that character-level models generally outperforms subword-level ones given enough model capacity.

\paragraph{Byte-Level Vocabularies}
The closest work to ours is the byte-level BPE vocabulary used in GPT-2, a large-scale English language model~\cite{radford:gpt2}. They however rely heavily on hard-coded merging rules and have not conducted any analysis on how their bye-level BPE impacts the quality of language modeling. A vocabulary consisting purely of bytes has previously been used in several natural language processing tasks: part-of-speech tagging and named entity recognition \cite{gillick:naacl2016}, translation\cite{costajussa:2017}, machine reading \cite{kenter2018byte} and speech recognition \cite{li:icassp2019}.


\paragraph{Transformer with Convolution or RNN}

There are evidences for performance gains from combining Transformer with convolutional or recurrent layers in the area of NMT~\cite{chen2018best},  speech recognition ~\cite{li:icassp2019,mohamed2019transformers} and language modeling ~\cite{wang2019language}.

\section{Conclusion}
We proposed BBPE which builds a byte-level subword vocabulary for machine translation. It results in a much more compact vocabulary than character-based ones do without the loss of performance. In multilingual settings, the former often outperforms the latter. BBPE does not have any out-of-vocabulary tokens, allowing us to transfer a model using BBPE between languages with non-overlapping vocabularies. This transfer learning paradigm is actually very generic and can be applied to any languages and datasets for performance gain or training acceleration. With the same vocabulary size, BBPE segments sentences into shorter sequences than character-based methods do, leading to faster training and inference. Our future work includes: eliminating source-target sentence length imbalance; evaluating BBPE in one-to-many and many-to-many translation settings; exploring better segmentation algorithms for byte-level subwords.

\bibliography{aaai20}
\bibliographystyle{aaai}

\end{document}